\documentclass{article}

\usepackage[final,sglblindworkshop]{neurips_2026}
\workshoptitle{Evaluation of Interactive Agents}

\usepackage[utf8]{inputenc}
\usepackage[T1]{fontenc}
\usepackage[hypertexnames=false]{hyperref}
\usepackage{xurl}
\usepackage{booktabs}
\usepackage{amsfonts}
\usepackage{amsmath}
\usepackage{microtype}
\usepackage{xcolor}
\usepackage{graphicx}
\usepackage{float}
\newcommand{\plotdep}[1]{#1}
\usepackage{twemojis}
\newcommand{\hficon}{\twemoji{1f917}}
\newcommand{\hfdataseturl}{https://huggingface.co/datasets/JetBrains-Research/SWE-TaskFlow-SWE-Bench-Pro}
\newcommand{\codeicon}{\twemoji{1f4bb}}
\newcommand{\codeurl}{https://github.com/JetBrains-Research/SWE-TaskFlow-Generation-Public}

\title{Opinion: Coding-Agent Benchmarks Should Match Their Users' Task Flows}

\author{
  Igor Slinko$^{1}$ \quad Yaroslav Golubev$^{1}$ \quad Sergey Titov$^{1}$ \\
  $^{1}$JetBrains Research
}

\begin{document}

\maketitle

\begin{abstract}
The evaluation of coding agents generally strives to be as realistic as possible. In our study, we collect 4,782 agent sessions of real software engineers in JetBrains IDEs, which we call \emph{Production Sessions}.
Since our subject is interactive agents, we study the sessions with at least three user messages (33\% of the sample). These long sessions differ from issue-derived benchmark tasks in two ways: (i) user requests span a far wider mix of task types---questions about the project's code, planning, review, refactoring, execution---and (ii) users switch between types throughout a session. Long-session samples from three public interaction corpora exhibit markedly different \emph{Task Flows} (the distributions of session lengths, task types, and type-to-type transitions), so no single interaction distribution is universally realistic: benchmarks should name a target use case and calibrate to measurements from it. We present SWE-TaskFlow, an approach for transforming any issue-derived benchmark: it preserves the verified tasks and tests while steering the interaction toward a target Task Flow through prompt splitting and verifiable repository QA, with a TaskFlow Alignment Score (TFAS) for selecting among generated trajectories. In a pilot on 700 SWE-Bench Pro tasks, \plotdep{solving the task sequentially in several steps approximately doubles agent cost without a stable change in resolve rate}: the interaction protocol itself is an important dimension of evaluation.
\end{abstract}

\begin{center}
  \refstepcounter{figure}
  \centering
  \vspace{-0.6em}
  \includegraphics[width=0.92\linewidth]{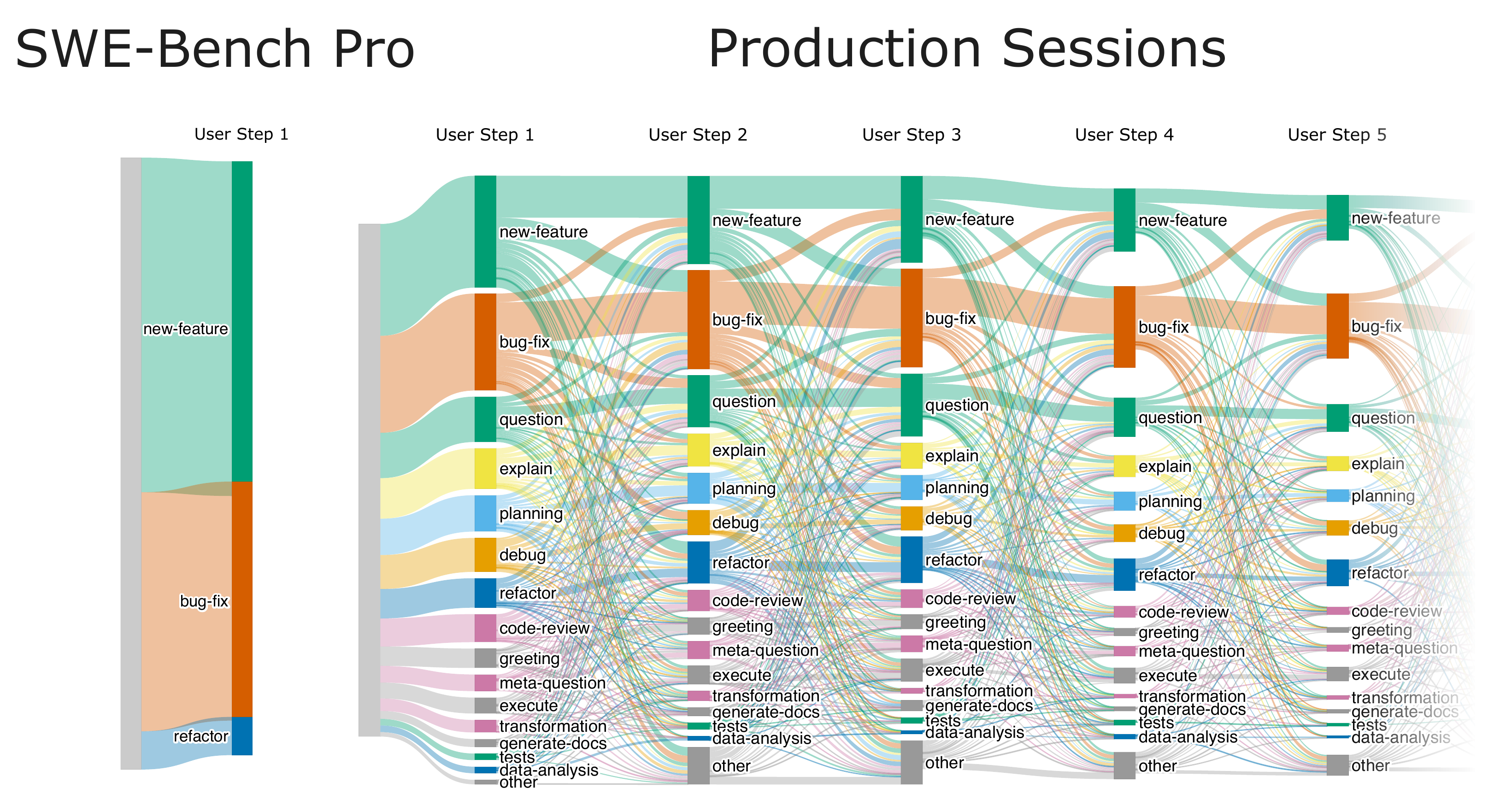}
  \vspace{-0.6em}

  {\small Figure~\thefigure: Left: an issue-derived benchmark (SWE-Bench Pro) presents each task as a single user turn, almost always a new feature or a bug fix. Right: Task Flow of \textit{Production Sessions} that contain at least three messages; nodes are primary intents and link widths are transition frequencies.}
  \label{fig:taskflows}
\end{center}

\section{Introduction}
The field of agent evaluation aims to assess agents in realistic scenarios. To see what such a scenario might look like, we start our study by getting access to 4,782 agent sessions of real software engineers working in JetBrains IDEs; we call them \textit{Production Sessions}. In this data, we focus on the long-interaction slice: sessions with at least three user messages (accounting for as much as 33\% of the sample). We describe such a corpus by its \emph{Task Flow}: session lengths, the task types of user messages, and the transitions between types from one message to the next (formal definition in Section~\ref{sec:users}). Two observations about the Task Flow of these sessions motivate this work (Figure~\ref{fig:taskflows}).

First, \emph{user prompts are far more varied than benchmark tasks}. In SWE-Bench Pro~\citep{deng2025swebenchpro} (left panel of Figure~\ref{fig:taskflows}), new features (54.7\%) and bug fixes (39.1\%) account for 93.8\% of tasks (labels from Section~\ref{sec:users}); in Production Sessions the same two intents cover well under half of user messages, and the remainder asks the agent to explain the project's code, answer questions, plan, review, refactor, and execute. Second, \emph{users switch task types throughout a session}: one that opens with a feature request often continues with a question about existing behavior, a planning step, and a review request. Issue-derived benchmarks ignore this: SWE-bench~\citep{jimenez2024swebench} and successors such as SWE-Bench Pro, SWE-rebench, and DeepSWE~\citep{deng2025swebenchpro,badertdinov2025swerebench,huang2026deepswe} deliver one detailed issue and score one final patch. Their infrastructure and tests are valuable, but their \emph{interaction layer}---the user messages the agent receives and the order in which it receives them---measures implementation from an upfront specification rather than adaptation to requirements unfolding across a session.

Which types of interaction steps to add depends on what users you target: we also applied our classifier to three public interaction corpora~\citep{baumann2026swechat,tang2026programming,dataclaw} and obtained visibly different Task Flows for each (Appendix~\ref{app:public-flows})---a distribution measured on one product does not transfer to another. Our position is therefore: \textbf{there is no universal realistic coding-agent benchmark; benchmark builders should state the target use case, measure its Task Flow, and calibrate the benchmark's interaction layer to it}.

\textbf{Contributions.} We contribute (i) a Task-Flow analysis of long Production Sessions that makes a measured interaction distribution the design target; (ii) SWE-TaskFlow, an approach that turns verified SWE tasks into replayable multi-turn trajectory candidates while preserving the original requirements and tests; and (iii) a per-turn evaluation design in which one task yields several verifiable signals instead of a single terminal score, with a pilot on 700 SWE-Bench Pro tasks showing that the interaction protocol is a real evaluation variable. We publicly release the \href{\codeurl}{\codeicon~\textbf{generation~code}} and all generated tasks with agent trajectories as one \href{\hfdataseturl}{\hficon~\textbf{public~dataset}} (details in Appendix~\ref{app:pipeline}).

\section{Task Flows of Production Sessions}
\label{sec:users}

\textbf{Data and classification.} A Production Session is one conversation between a developer and the coding agent inside the IDE. We analyze only the user's side of it: %
roughly 14,000 user messages across the 4,782 sessions, collected from users of JetBrains IDEs in the course of their everyday work. Each message is independently classified by an open-weight LLM into one of 16 task types, or intents (bug fix, new feature, explain, planning, code review, execution, and so on); the definitions were iterated with manual review, and three classifier models agree on 76.1\% of labels (taxonomy, prompt, and robustness check in Appendix~\ref{app:taxonomy}). The same pipeline is applied to Production Sessions, public corpora, and benchmark prompts, so every distribution shares the same footing.

\textbf{Task Flow.} Formally, a corpus-level \emph{Task Flow} is a triple of distributions $(P_L, P_T, P_R)$: (i) session lengths, (ii) task types over messages, and (iii) directed transitions between adjacent messages. The transition distribution preserves ordering information that a label histogram alone would lose.

\textbf{Production Sessions.} Of the 4,782 sessions (analyzed only as anonymized aggregate statistics), 50.9\% contain a single user message and 33\% at least three; the 75th and 90th percentiles are three and eight messages (Appendix~\ref{app:taxonomy}, Figure~\ref{fig:empirical}), and first messages average 30.7 words against 16--19 at later positions. Bug fixes (25.9\%) and new features (18.9\%) lead but do not dominate as they do in issue-derived datasets. All subsequent Task-Flow analysis conditions on this long-session slice.

\textbf{QA on project code is a frequent intent.} In long Production Sessions, \textsc{explain} requests (questions about how the project's code works) account for 6.1\%, while general programming \textsc{question}s for another 9.7\%. Explanation is the intent that issue-derived benchmarks lack; it remains common at every position (Figure~\ref{fig:taskflows}) and motivates the QA augmentation of Section~\ref{sec:taskflow}.

\textbf{Public corpora have different Task Flows.} We applied the same pipeline to long-session samples from SWE-chat, SpecStory, and DataClaw~\citep{baumann2026swechat,tang2026programming,dataclaw}. Their Task Flows visibly differ from Production Sessions and from each other in both intents and transitions (Appendix~\ref{app:public-flows}): there is no ground-truth interaction distribution, and the target must be measured in its own setting.

\section{SWE-TaskFlow: Transforming Verified Tasks}
\label{sec:taskflow}

Building a new interactive benchmark for every target is expensive and would discard the investment behind existing verified SWE tasks (repository states, execution environments, test-based verifiers). SWE-TaskFlow is therefore a transformation of existing issue-derived datasets, instantiated here on SWE-Bench Pro: the repository, requested change, environment, and final tests remain fixed, while the interaction layer is regenerated. Related transformations (Appendix~\ref{app:related}) also fix trajectories before inference for exact replayability; SWE-TaskFlow differs in calibrating their shape to a measured Task Flow rather than choosing it ad hoc.

\textbf{Dataset augmentation.} We propose two augmentations to the existing tasks. The first, \emph{prompt splitting}, targets overspecification: a SWE-bench-style issue states the goal, context, constraints, and often the approach in one polished message, whereas real users reveal requirements gradually (Section~\ref{sec:users}). Splitting turns one issue into $K$ ordered, self-contained user turns ($K{=}3$ here; adjacent parts can be merged). The LLM-based splitter must preserve every original requirement, introduce none, keep technical artifacts verbatim, and make each turn actionable in the repository state left by earlier turns. Issues that do not pass these checks stay single-turn; 700 of the 731 public SWE-Bench Pro tasks are retained after splitting and evaluation-context checks. The second, \emph{repository QA}, generates zero to three verifiable questions about repository behavior, injecting the \textit{explain} intent that benchmarks lack. Question placement---before, between, or after issue turns---determines which transitions appear; together the augmentations yield one to six user turns per task, of which candidates with at least three form the long-session slice.

\textbf{Where the questions come from.} Questions are mined from an agent trajectory that solved the original issue, read together with the live repository state. Every candidate carries a hidden reference answer and must be answerable from the repository, stay valid before and after the issue patch, not leak the solution, and pass an executable \emph{proof script} run in a clean repository copy at creation time. At evaluation time the agent answers in text and is judged against the reference (full generation methodology and pipeline in Appendix~\ref{app:pipeline}).

\textbf{Measuring alignment.} To make ``closer to the target Task Flow'' explicit and optimizable, we propose the \emph{TaskFlow Alignment Score (TFAS)}. Let $P_k$ and $Q_k$ be target and benchmark distributions for $k\in\{L,T,R\}$: session-length bins, task types, and directed transitions between adjacent messages. TFAS uses 13 of the 16 classes---\textsc{greeting}, \textsc{meta-question}, and \textsc{other} are not instantiable as repository tasks; transitions are computed over retained messages. With base-2 Jensen--Shannon divergence, we set $S_k = 1-\operatorname{JSD}_2(P_k,Q_k)$ and $\operatorname{TFAS}=100\cdot\left(S_L \cdot S_T \cdot S_R\right)^{1/3}$, where each $S_k$ lies in $[0,1]$, so TFAS lies in $[0,100]$, with 100 an exact match of all three distributions. Candidate selection maximizes TFAS over the generated family subject to task-count and diversity constraints, and each component must accompany the aggregate (computation details, the optimized selection, and ablations in Appendix~\ref{app:tfas}).

\section{Pilot: The Interaction Layer Changes the Measurement}

\textbf{Does progressive disclosure change difficulty?} With \textit{mini-swe-agent}~\citep{yang2024sweagent} as a common scaffold, we evaluated \textit{Single} and \textit{Split} prompts on the same 700 tasks with four models. \plotdep{\textit{Split} approximately doubled agent cost} (Appendix~\ref{app:experiments}, Figure~\ref{fig:interaction-tax}). \plotdep{Resolve-rate shifts were mixed across models and largely comparable to run-to-run variation, so these runs establish no success ordering but show that progressive disclosure reliably changes the effort an agent expends.}

\begin{figure}[t]
  \centering
  \includegraphics[width=0.8\linewidth]{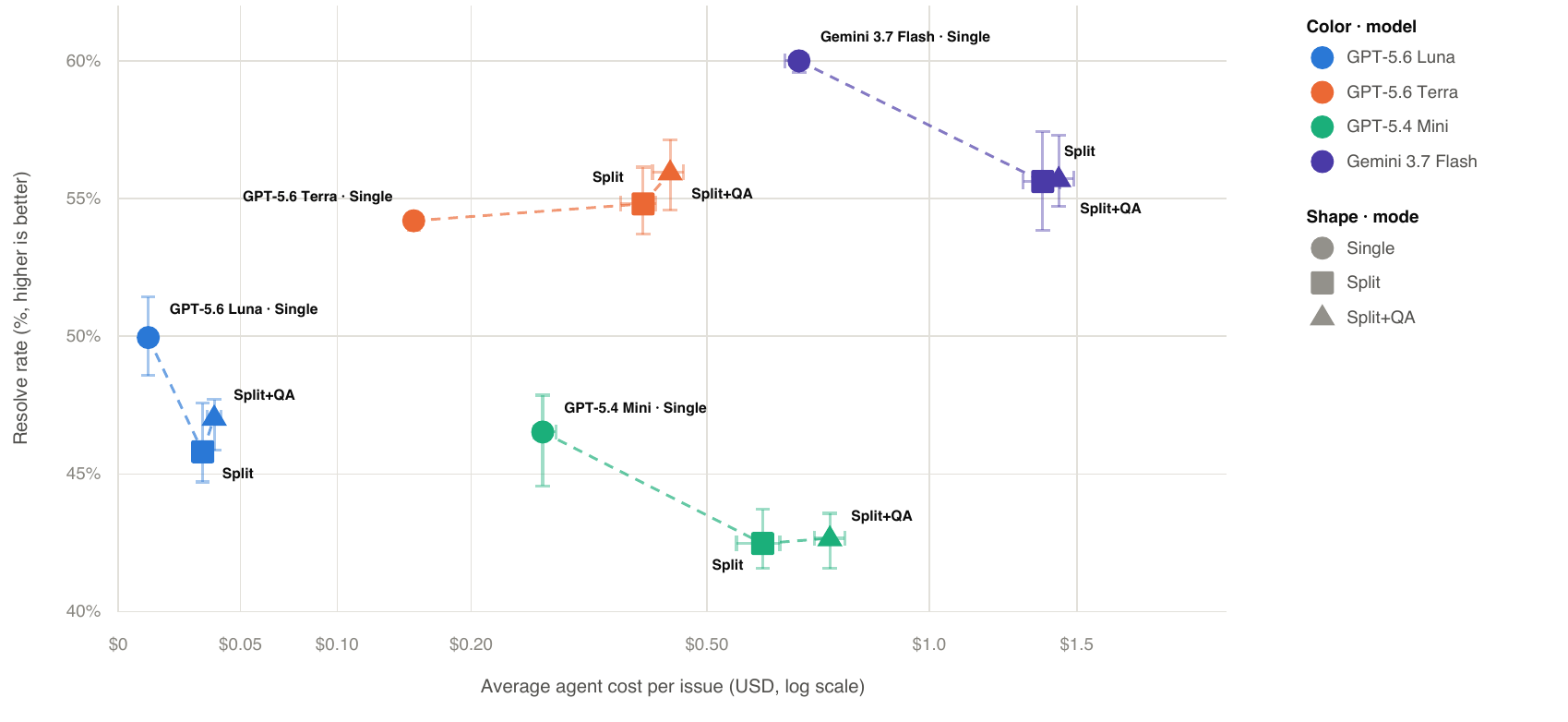}
  \vspace{-0.5em}
  \caption[Resolve rate versus average agent cost per issue]{Resolve rate versus mean agent cost per issue (log scale) on the 700 tasks; markers are three-run means, whiskers min--max (not confidence intervals).}
  \label{fig:pilot-metrics}
  \vspace{-0.5em}
\end{figure}

\textbf{Calibrated QA turns.} Tuning the number and positions of QA turns to maximize TFAS against the long-session Production target (\emph{Split+QA} in Figure~\ref{fig:pilot-metrics}) raises TFAS from 59.6 for pure \textit{Split} to 77.8 (Appendix~\ref{app:tfas}, Table~\ref{tab:tfas-variants}; Task Flow in Figure~\ref{fig:overview}). \plotdep{The calibrated QA turns add moderate cost on top of \textit{Split}; resolve rate again shows no stable ranking across models} (Figure~\ref{fig:pilot-metrics}), \plotdep{and QA order has no stable effect on resolve rate or answer accuracy} (Appendix~\ref{app:experiments}). \plotdep{Answer accuracy on the selected benchmark spans 69.8--83.0\% across the four models}, a second verified signal alongside resolve rate (Appendix~\ref{app:qa-accuracy}).

\textbf{Per-turn verified signals.} A \textit{Single} trajectory produces one binary signal, while a \textit{Split+QA} trajectory yields a test-suite verdict after each issue turn (scored only against requirements revealed so far) and a judged answer per QA turn (Appendix~\ref{app:experiments}, Figure~\ref{fig:checkpoint-signals}). Denser verified signals support finer-grained diagnosis and ranking-based training~\citep{choi2026gopo,yari2026amir,yu2026pairgrpo}.

\section{Discussion: Toward Richer Task Flows}
\label{sec:discussion}

\textbf{Most user intents still lack augmentations.} SWE-TaskFlow currently covers prompt splitting for implementation intents and repository QA for \textsc{explain} (cf.\ SWE-QA, SWE Atlas, RepoProbe~\citep{peng2026sweqa,raghavendra2026atlas,yang2026repoprobe}). Everything else in Figure~\ref{fig:taskflows} (general questions, planning, code review, refactoring, execution, debugging without a fix, etc.) still needs its own augmentation: a way to generate turns of that intent on a verified task and to check the response. Most are open problems: it is unclear how to verify a planning or code-review turn automatically. A measured Task Flow tells us which augmentations are missing and in what proportion.

\textbf{Static and reactive interaction are complementary regimes.} Fixing all user messages before inference enables exact replay and paired comparisons: SWE-TaskFlow measures progressive disclosure, memory, and adaptation, but not feedback reacting to what the agent did. The reactive benchmarks in Appendix~\ref{app:related} (withheld requirements, simulated users, injected edits, help-seeking) measure exactly that, at the price of simulator assumptions and trajectory variance. A full evaluation suite can combine a calibrated static core with targeted reactive slices.

\textbf{Limitations.} Long Production Sessions define only one use case, though the approach applies to any measured target (Appendix~\ref{app:recipe}). The intent classifier lacks a human-labeled accuracy set (Appendix~\ref{app:taxonomy}: inter-model agreement 76.1\%, manually checked samples per intent). Some issues resist splitting, and QA difficulty is generator-dependent. Finally, each pilot condition ran three times with min--max ranges, not confidence intervals; no comparison carries a significance test, so resolve-rate readings are directional only.

\section{Conclusion}
\label{sec:conclusion}

An interactive benchmark must state whose interaction distribution it represents: real users mix and switch task types in ways issue-derived benchmarks miss, and public corpora do not represent each other. SWE-TaskFlow addresses this gap while preserving verified tasks: splitting, verifiable QA, and TFAS-selected placement turn issue-derived benchmarks into calibrated, replayable multi-turn evaluations.

\bibliographystyle{plainnat}
\bibliography{references}

\clearpage
\appendix

\section{Task Taxonomy and Classification}
\label{app:taxonomy}

This appendix documents how every Task Flow in the paper is produced: the classification pipeline, the full 16-label taxonomy with a prompt excerpt, an inter-model robustness check, and summary statistics of the classified sessions.

\textbf{Classification pipeline.}
Each user message is classified independently by an open-weight LLM (Gemma 4-31B) that receives the message text together with the label definitions below and returns exactly one \texttt{task\_type} as structured output; when a message contains several requests, the classifier assigns the dominant actionable intent. The same pipeline, prompt, and definitions are applied to Production Sessions, the three public interaction corpora, and benchmark prompts, so every Task Flow in the paper is computed on the same footing. The taxonomy itself is grounded in industrial practice: it originates from an intent categorization used at JetBrains to analyze production assistant usage, which we consolidated into the 16 labels below. We chose it over published taxonomies because it was iterated directly against the kind of IDE sessions we study, and because it keeps IDE-specific intents---execution, meta-questions about the assistant itself---separable from repository tasks. During prompt development, the authors manually reviewed samples of classifications, revised ambiguous label instructions, and re-ran the pipeline; a quantitative inter-model robustness check is reported below.

\begin{description}
  \item[tests] Write or extend test coverage. Fixing an existing failing test belongs to bug-fix or debug.
  \item[bug-fix] Fix an existing error, bug, or exception. A bare stack trace defaults to a fix.
  \item[new-feature] Add new functionality.
  \item[generate-docs] Write documentation, comments, or README files.
  \item[refactor] Improve same-language code structure without changing behavior.
  \item[transformation] Convert code between formats, languages, library versions, or paradigms.
  \item[debug] Investigate why something fails without fixing it.
  \item[planning] Plan a change, design an approach, or evaluate a plan or specification.
  \item[explain] Explain how specific code in the user's project works.
  \item[question] Ask a general programming, framework, library, or tool question.
  \item[meta-question] Ask about the assistant, IDE, quota, billing, or environment.
  \item[code-review] Review specific changes for style, correctness, or performance.
  \item[execute] Run a command, build, test, query, deployment, or external action.
  \item[data-analysis] Analyze data or build and interpret queries, charts, or metrics.
  \item[greeting] Provide no programming content beyond a greeting, ping, or meaningless input.
  \item[other] Provide residual non-programming content such as thanks or feedback.
\end{description}

\textbf{Prompt excerpt.} The classifier receives one user message at a time. The task-type portion of the prompt is:
\begin{quote}
\small
Classify the primary intent of a developer message from a coding-assistant session. Assign exactly one \texttt{task\_type}. Focus on what the user asks the assistant to do, not on the topic, language, or solution. If the message contains several requests, choose the dominant actionable intent. Use only one label from the supplied taxonomy and return the requested structured output.
\end{quote}

\textbf{Inter-model robustness.} To estimate how much the labels depend on the choice of classifier, we classified the same 100 sessions with three open-weight models: Qwen 3.6-35B, Gemma 4-31B, and GPT-OSS 20B. Pairwise exact agreement on \texttt{task\_type} was 79.2\%, 73.7\%, and 75.5\%, respectively, for a mean of 76.1\% (Table~\ref{tab:classifier-agreement}). This measures inter-model robustness rather than accuracy: no formal human-labeled test set is available.

\begin{table}[t]
  \centering
  \caption{Exact agreement on \texttt{task\_type} over the same 100-session sample.}
  \label{tab:classifier-agreement}
  \begin{tabular}{lr}
    \toprule
    Model pair & Agreement \\
    \midrule
    Qwen 3.6-35B / Gemma 4-31B & 79.2\% \\
    Qwen 3.6-35B / GPT-OSS 20B & 73.7\% \\
    Gemma 4-31B / GPT-OSS 20B & 75.5\% \\
    \midrule
    Mean pairwise agreement & 76.1\% \\
    \bottomrule
  \end{tabular}
\end{table}

\begin{figure}[t]
  \centering
  \includegraphics[width=0.62\linewidth]{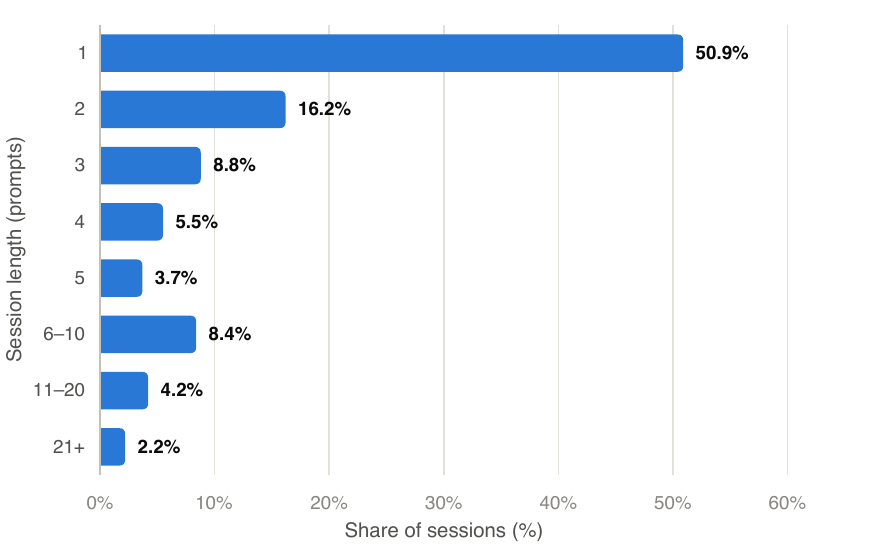}
  \caption{Production Sessions by number of user messages. The long-session slice studied throughout the paper aggregates all bins with at least three messages.}
  \label{fig:empirical}
\end{figure}

\textbf{Session statistics.} Figure~\ref{fig:empirical} summarizes the classified sample: the distribution of Production Sessions by number of user messages.

\section{Task Flows in Public Interaction Corpora}
\label{app:public-flows}

\begin{figure}[p]
  \centering
\includegraphics[width=0.70\linewidth]{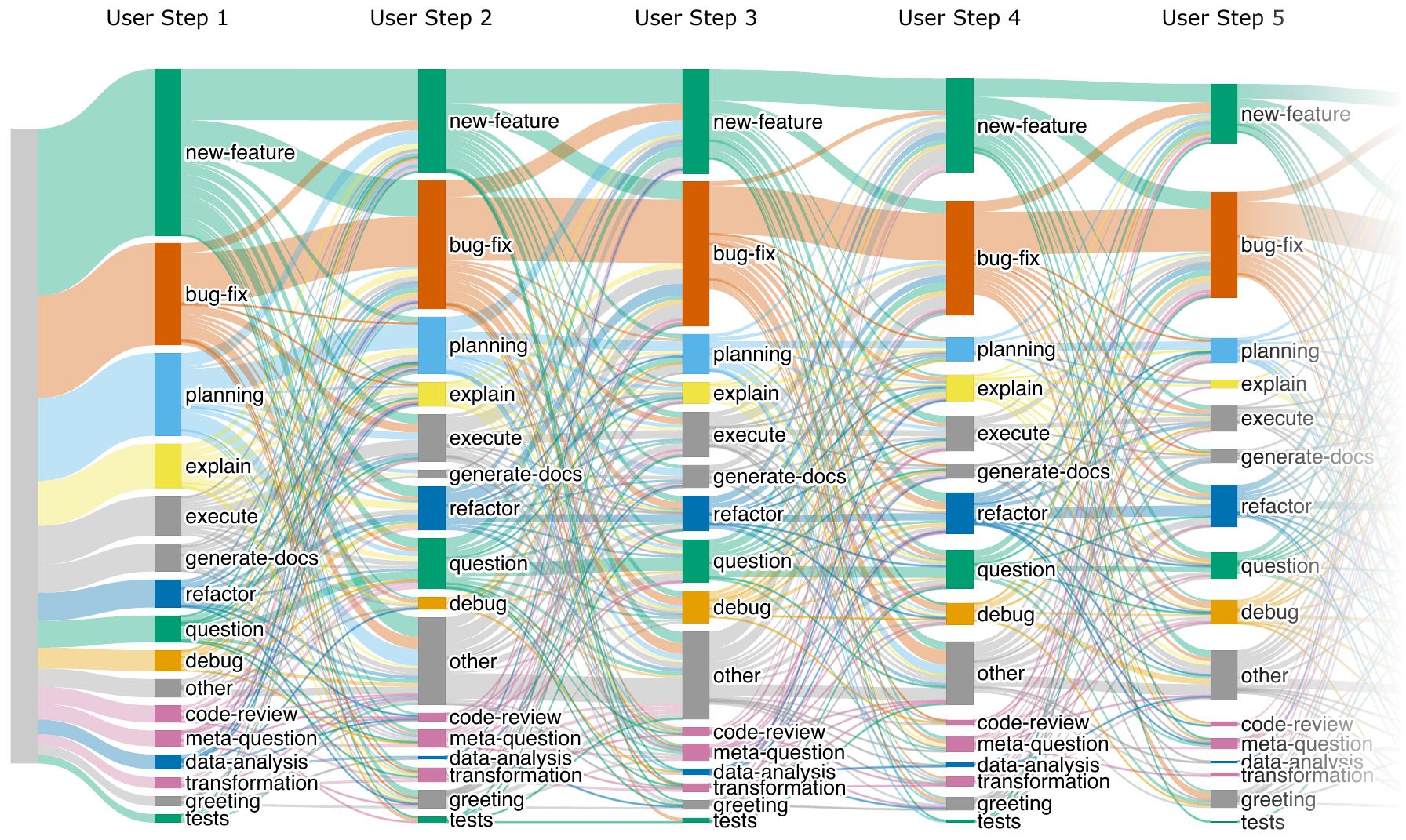}
  \caption{SpecStory long-session Task Flow: first five user messages of the 521 of 1,000 classified sessions with at least three; full 16-label taxonomy.}
  \label{fig:specstory-flow}
\end{figure}

\begin{figure}[p]
  \centering
\includegraphics[width=0.70\linewidth]{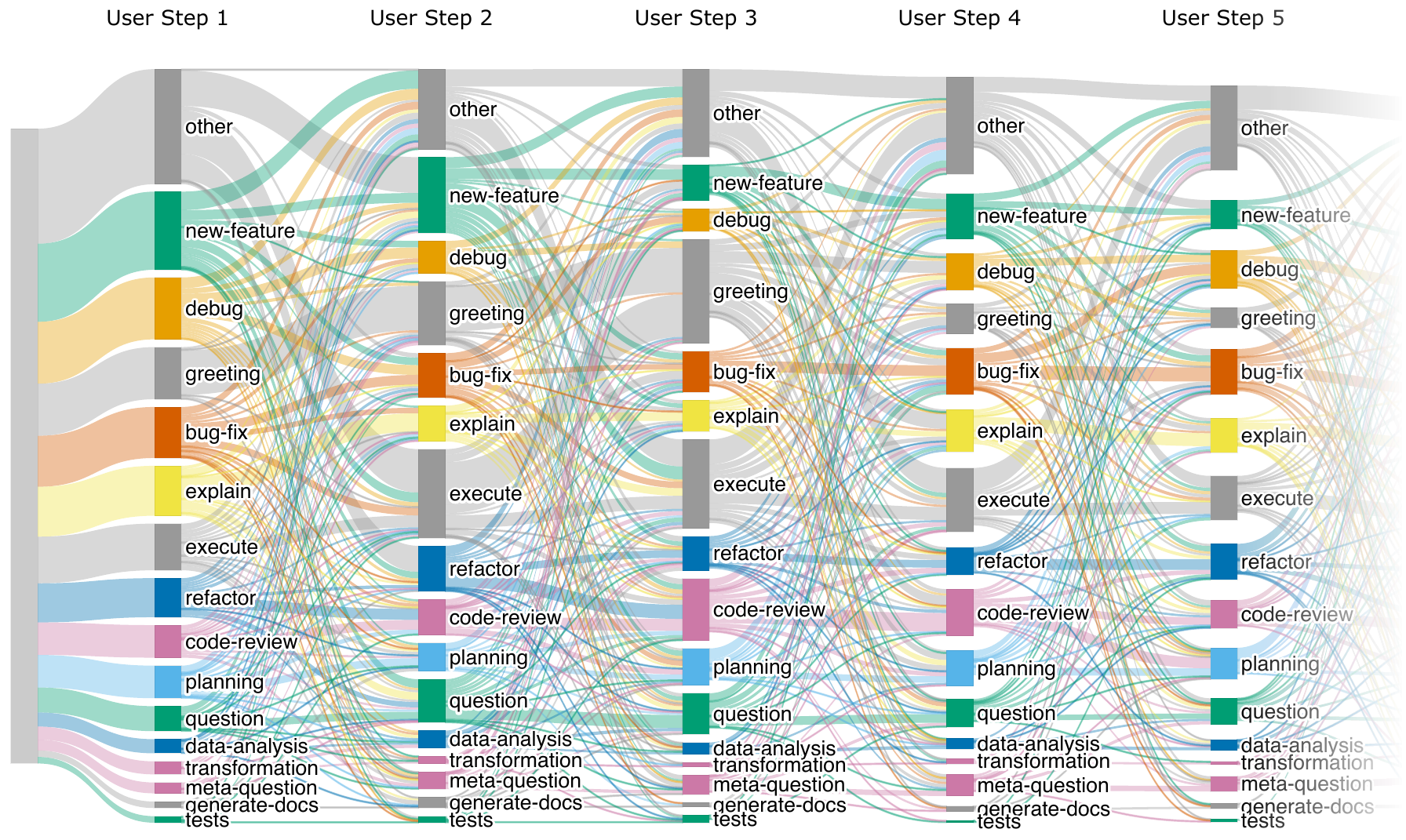}
  \caption{DataClaw long-session Task Flow: first five user messages of the 585 of 1,000 classified sessions with at least three; full 16-label taxonomy.}
  \label{fig:dataclaw-flow}
\end{figure}

\begin{figure}[p]
  \centering
\includegraphics[width=0.70\linewidth]{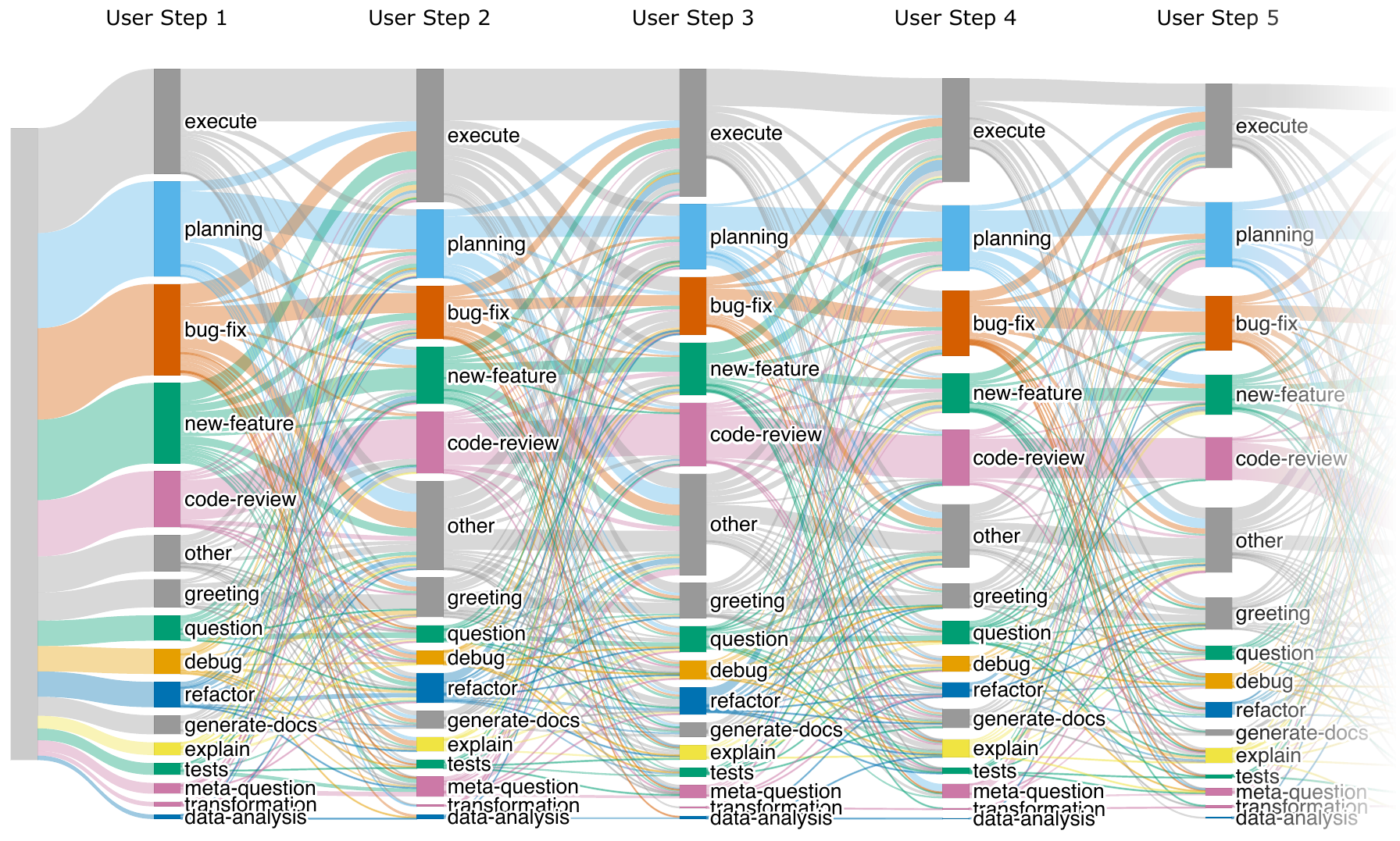}
  \caption{SWE-chat long-session Task Flow: first five user messages of the 698 of 1,000 classified sessions with at least three; full 16-label taxonomy.}
  \label{fig:swe-chat-flow}
\end{figure}

We applied the classification pipeline of Appendix~\ref{app:taxonomy} to long-session samples (at least three user messages) from three public interaction corpora, 1,000 classified sessions each. \textbf{SpecStory} (Figure~\ref{fig:specstory-flow}) consists of public IDE-native session exports analyzed by Programming by Chat~\citep{tang2026programming}: of the three corpora its Task Flow sits closest to Production Sessions, but with a larger planning share. \textbf{DataClaw} (Figure~\ref{fig:dataclaw-flow}) collects community-shared coding-agent session exports~\citep{dataclaw}: it is dominated by execution and short operational turns. \textbf{SWE-chat} (Figure~\ref{fig:swe-chat-flow}) is an interaction corpus from a chat-based coding setting~\citep{baumann2026swechat}: execution and planning dominate its flows. These corpora are comparison datasets; the paper's calibration target remains Production Sessions. Their collection mechanisms, user populations, and products all differ, so the figures support a qualitative claim: the interaction distribution depends on the source, and none of them is ground truth. Large-scale behavioral analysis of coding sessions has precedent---Programming by Chat characterizes 11,579 real-world AI-assisted IDE sessions~\citep{tang2026programming}; what our Task-Flow analysis adds is the transition view and its use as a calibration target for benchmarks.

\clearpage
\section{Calibrating to a New Target}
\label{app:recipe}

This appendix turns the calibration approach of Sections~\ref{sec:users}--\ref{sec:taskflow} into a step-by-step procedure for a new target. For each step we state what goes in and what comes out, how we instantiated it in this paper, and what a benchmark builder working in a different domain has to decide anew. The procedure assumes a source benchmark with verified tasks (repository state, environment, tests) whose interaction layer is to be re-shaped; it does not assume that the source benchmark and the target share a domain.

\begin{enumerate}
  \item \textbf{Name the target.} A target is a user population, a product surface, a session filter, and a measurement window. \emph{Ours:} users of JetBrains IDEs during their everyday work, sessions with at least three user messages, one sample of 4,782 sessions. The filter is part of the definition: the same sample with a filter of at least one message is dominated by single-message sessions (50.9\%) and yields a different Task Flow. Task Flows also drift as products and models change, so a target should carry its measurement date, and a benchmark calibrated to it should state that date. \emph{New domain:} the population and filter should match the question the benchmark is meant to answer (for example, terminal-agent users with at least one tool call, or data-science notebook sessions longer than ten cells). Borrowing a target from a different product is not a shortcut: the three public corpora in Appendix~\ref{app:public-flows} differ from each other and from Production Sessions.
  \item \textbf{Sample user-side messages.} Only the user's messages are needed---no agent output, no code, no tool logs---which keeps collection cheap and limits privacy exposure. \emph{Ours:} roughly 14,000 user messages across 4,782 sessions, analyzed in aggregate only. \emph{New domain:} the sample must be large enough for the transition distribution to be stable; with 13 retained labels plus \textsc{start}/\textsc{end}, the transition support has about 200 cells, most of them rare. Resample complete sessions with the bootstrap and check that the component scores $S_k$ of Appendix~\ref{app:tfas} are stable before fixing the target; if they are not, enlarge the sample or coarsen the taxonomy.
  \item \textbf{Classify with a fixed taxonomy.} Every message receives exactly one label from a taxonomy whose definitions are frozen before measurement. \emph{Ours:} the 16-label taxonomy, prompt, and open-weight classifier of Appendix~\ref{app:taxonomy}, with a 76.1\% mean pairwise inter-model agreement on 100 sessions. \emph{New domain:} reuse the taxonomy where possible and extend rather than replace it, so that Task Flows measured for different targets remain comparable; every new label must be marked as instantiable or not as a repository task, because only instantiable labels enter TFAS. Run the inter-model check on at least 100 sessions before measuring; low agreement signals ambiguous label definitions, which should be revised and re-run before the target is fixed, not corrected afterwards.
  \item \textbf{Measure the target Task Flow.} Compute $(P_L, P_T, P_R)$ exactly as in Appendix~\ref{app:tfas}: length bins, task types over the retained labels, and directed transitions with explicit \textsc{start}/\textsc{end} edges after filtering non-instantiable messages. \emph{Ours:} the long-session Production target (Figure~\ref{fig:taskflows}, right). \emph{New domain:} publish the three vectors together with the benchmark. They are the calibration target, they let others recompute TFAS and its components, and they are aggregate statistics that reveal nothing about individual sessions.
  \item \textbf{Find the gaps.} Classify the source benchmark's prompts with the same pipeline to obtain its own Task Flow $(Q_L, Q_T, Q_R)$ and compare it with the target component by component. Every task type that is frequent in the target but rare or absent in the benchmark, and every session length that occurs in the target but that the benchmark cannot produce, needs an augmentation that stays verifiable. \emph{Ours:} SWE-Bench Pro is single-turn and 93.8\% new-feature or bug-fix, so the length gap is closed by prompt splitting and the \textsc{explain} gap by repository QA; the remaining gaps---\textsc{question}, \textsc{planning}, \textsc{code-review}, \textsc{refactor}, \textsc{execute}, \textsc{debug}---are listed as uncovered in Section~\ref{sec:discussion}. \emph{New domain:} the gap table is the to-do list of augmentations, ordered by how frequent each missing intent is in the target. An intent without a verifier should be reported as uncovered rather than approximated by an unverifiable turn.
  \item \textbf{Generate candidate families and select.} For each verified task, enumerate the admissible combinations of augmentations (here: merge level of the split turns, number of QA turns, their placement), then choose one candidate per task to maximize TFAS against the target, subject to task-count and diversity constraints. \emph{Ours:} the randomized local search of Appendix~\ref{app:tfas}, with the component scores reported alongside the aggregate (Table~\ref{tab:tfas-variants}). \emph{New domain:} before granting the optimizer freedom along some dimension, check that moving along it does not change task difficulty; our QA-placement ablation (Appendix~\ref{app:experiments}) is this check for placement. Without it, a higher TFAS could be bought by changing what the benchmark measures.
  \item \textbf{Separate stress tests.} Any selection that maximizes alignment with observed frequencies pushes rare intents toward zero. Intents that are rare in the target but consequential for the product---destructive \textsc{execute} commands, security-relevant \textsc{code-review}---should be evaluated as explicit stress-test slices with their own task counts and reported separately from the calibrated core. \emph{Ours:} the released dataset keeps every experimental condition as a separate split, and the SWE-chat ablation (Table~\ref{tab:tfas-swe-chat}) is reported apart from the Production-calibrated selection. \emph{New domain:} decide which rare intents matter before optimizing, so that their slices are built deliberately rather than recovered from what the optimizer discarded.
\end{enumerate}

If no production data is available, the closest public interaction corpus (Appendix~\ref{app:public-flows}) can serve as a provisional target, provided the substitution is stated explicitly; TFAS still yields a measurable distance, and the target can be replaced once production measurements exist.

\section{Related Transformations of Verified Tasks}
\label{app:related}

SWE-TaskFlow builds on a growing family of works that add an interaction layer on top of existing verified tasks rather than collecting new ones. We group them into three families.

\textbf{Static transformations} fix the user messages before inference, as we do. Sharded decomposition of a task into a fixed sequence of prompts has been studied for HumanEval and LiveCodeBench, where progressive disclosure consistently changed measured performance~\citep{laban2025lost}, and Saving SWE-Bench lowers the specificity of the issue text while keeping the interaction single-turn~\citep{garg2025saving}. These transformations choose their trajectory shapes ad hoc; none calibrates the resulting interaction distribution to a measured user population, which is the gap SWE-TaskFlow targets.

\textbf{Reactive designs} make the environment respond to the agent instead of fixing the messages upfront. Ambig-SWE and Dialogue SWE-Bench withhold requirements and reveal them when the agent asks~\citep{vijayvargiya2026ambigswe,king2026dialogue}; SWE-Interact and SWE-Together employ LLM-simulated users that answer clarification questions~\citep{raghavendra2026interact,wu2026swetogether}; SWE-Touch injects user edits into the repository mid-trajectory~\citep{tan2026swetouch}; and HiL-Bench studies when agents seek human help~\citep{trinh2026hilbench}. Reactive designs measure adaptation to feedback but give up exact replayability: two runs of the same task can see different user messages (Section~\ref{sec:discussion} contrasts the two regimes).

\textbf{Repository QA} benchmarks evaluate question answering about a codebase as a task in its own right: SWE-QA, SWE Atlas, and RepoProbe offer complementary designs~\citep{peng2026sweqa,raghavendra2026atlas,yang2026repoprobe}. SWE-TaskFlow uses repository QA differently---as inserted turns inside a longer trajectory, with creation-time proof scripts guaranteeing that each question is answerable from the repository.

\section{Generation and Verification Pipeline}
\label{app:pipeline}

\begin{figure}[p]
  \centering
  \makebox[\linewidth][c]{\includegraphics[width=1.08\linewidth]{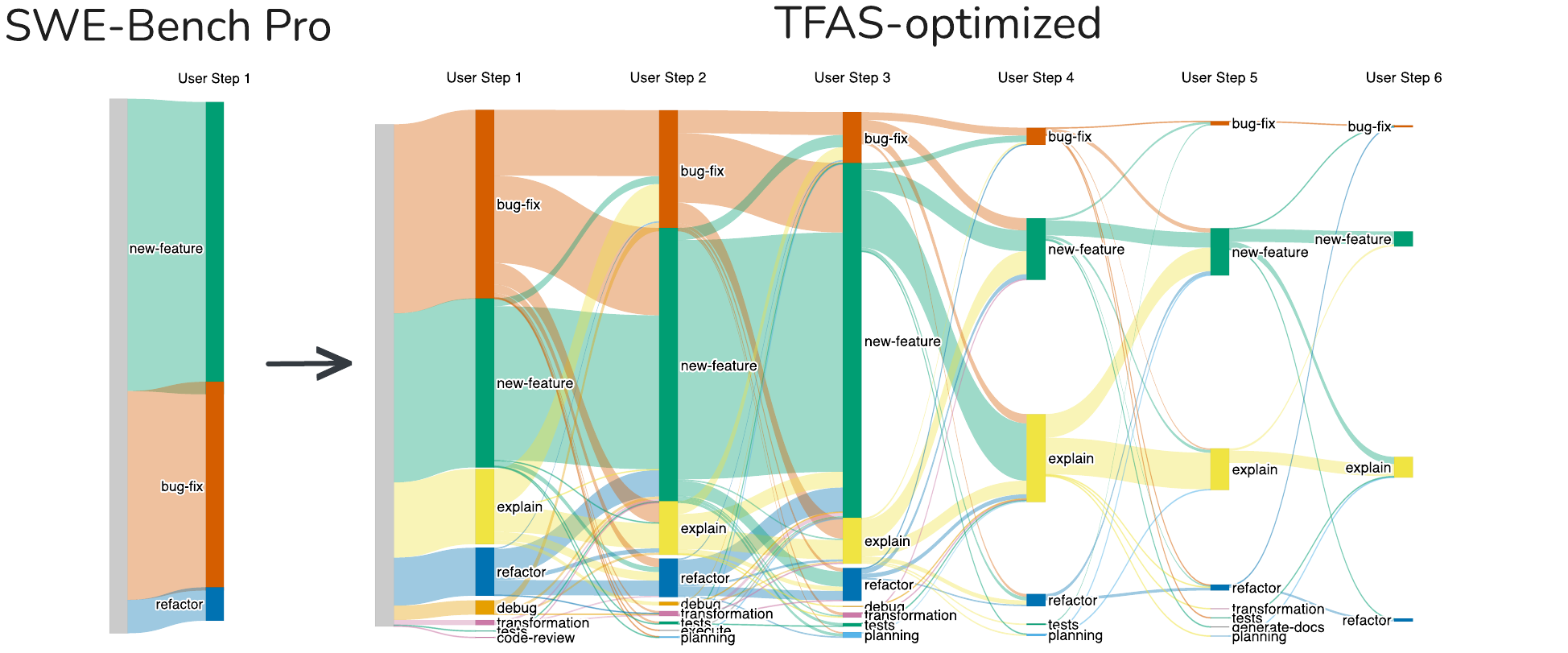}}
  \caption{Task Flow of SWE-Bench Pro (left) and of the TFAS-optimized selection (right)---the configuration reported in Table~\ref{tab:tfas-variants} with TFAS 77.8: one user prompt per issue becomes multiple sequential prompts spanning different task types. Repositories, requested changes, and final test-based verifiers are unchanged.}
  \label{fig:overview}
\end{figure}

\begin{figure}[p]
  \centering
  \includegraphics[width=\linewidth]{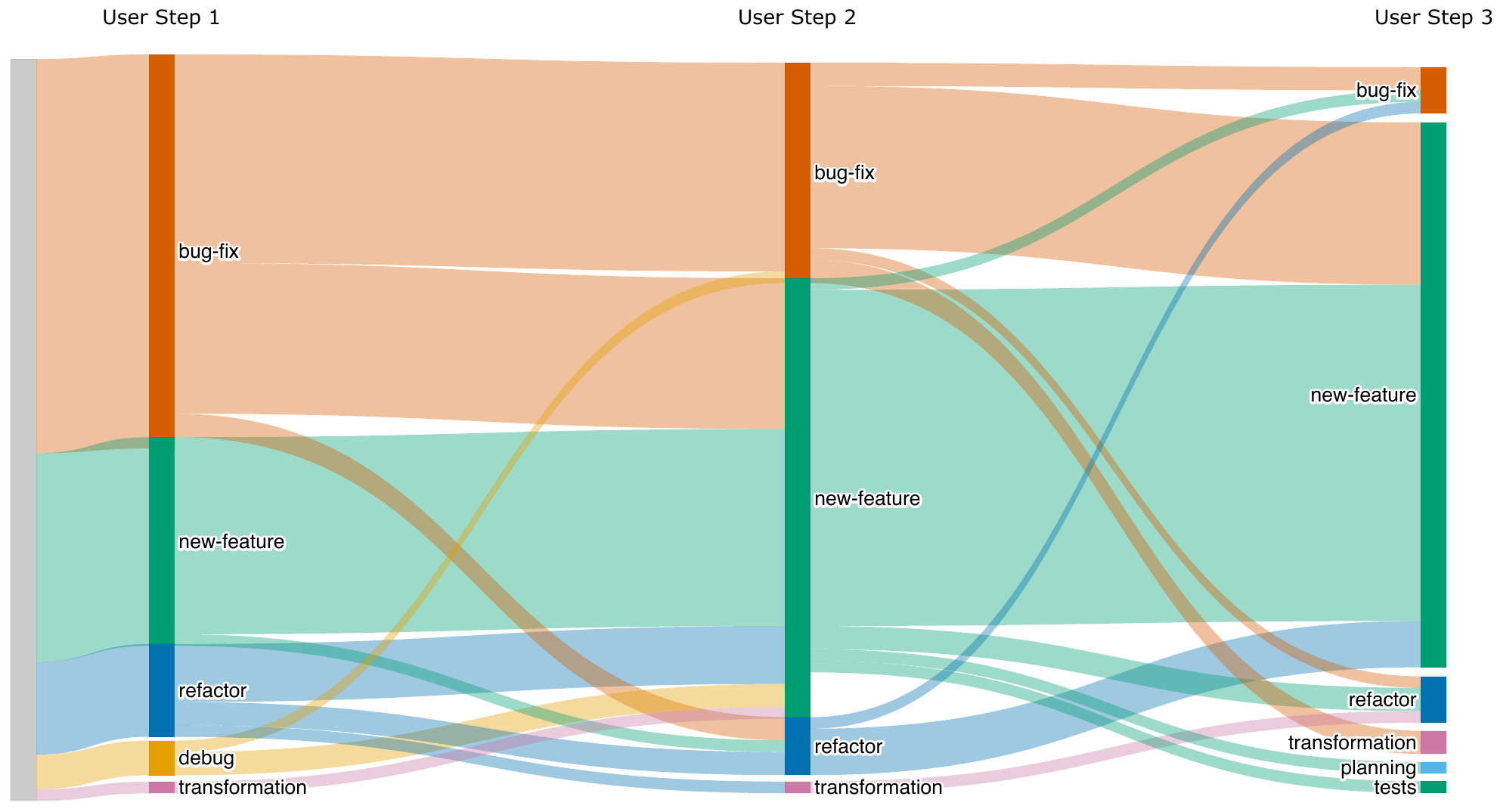}
  \caption{Task Flow after three-way prompt splitting. Splitting increases trajectory length and exposes several intents that were implicit in the original issue.}
  \label{fig:split-flow}
\end{figure}

\begin{figure}[t]
  \centering
  \includegraphics[width=\linewidth]{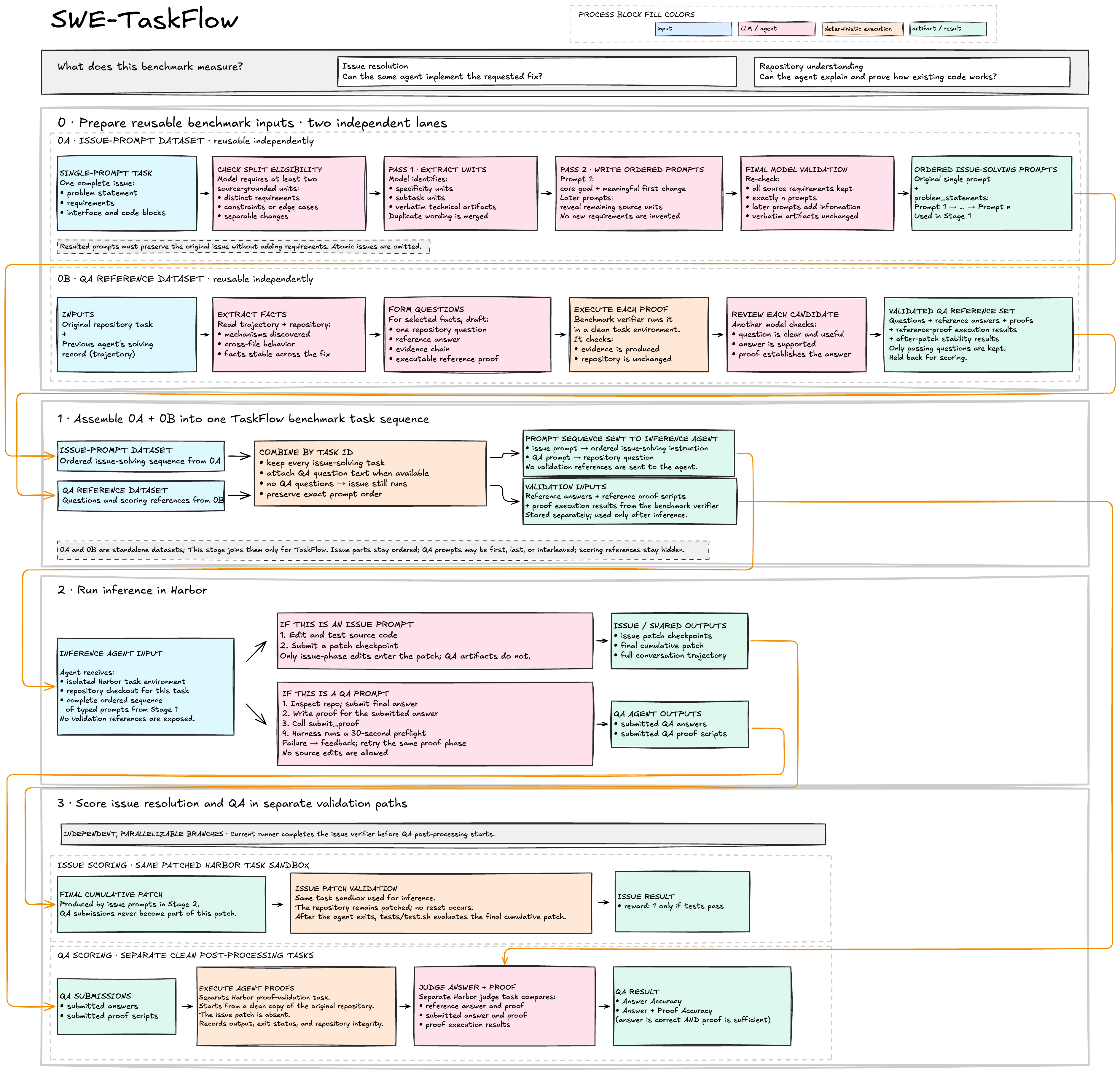}
  \caption{Complete generation, assembly, inference, and verification pipeline. Reference answers and proof artifacts remain hidden from the evaluated agent.}
  \label{fig:pipeline}
\end{figure}

Figure~\ref{fig:overview} contrasts the interaction layer of the source benchmark with the result of the transformation at the benchmark level, Figure~\ref{fig:split-flow} shows the intermediate Task Flow of pure \textit{Split} trajectories before QA insertion, and Figure~\ref{fig:pipeline} shows the complete generation, assembly, inference, and verification pipeline.

\textbf{Prompt-splitting contract.} The splitter receives an issue description and first decides whether three meaningful, ordered user turns are possible. It then extracts specificity units, subtask units, and verbatim technical artifacts before writing the sequence and performing a final model-based completeness check. An eligible decomposition must preserve every original requirement, introduce no new requirement, keep technical artifacts unchanged, avoid references to the decomposition process, and make each turn actionable in the repository state produced by earlier turns. Ineligible atomic issues remain single-turn. Figure~\ref{fig:split-flow} shows the Task Flow of the resulting \textit{Split} trajectories.

\textbf{Question-generation contract.} The generator reads a prior solving trajectory and the corresponding live repository state, extracts stable cross-file behavior, and proposes one to three questions with reference answers and evidence chains. Questions must be answerable from the repository, remain valid before and after the issue patch, and avoid revealing the issue solution. In the current pipeline, generation runs in Codex using Sol or Terra: the TFAS-optimized selection and the \textit{Split+QA} results use the Sol-generated pool---454 questions across 255 of the 700 tasks (Appendix~\ref{app:tfas})---while the 64-task placement ablation uses Terra-generated questions (Appendix~\ref{app:experiments}). A separate model pass reviews clarity and repository support. Executable proof scripts serve as mandatory creation-time instrumentation: for each candidate question, the pipeline generates a reference script, executes it in a clean environment, checks that it produces evidence without modifying the repository, and retains only candidates whose proof passes. This requirement bounds the admissible question space---questions without observable evidence, such as the rationale for an architectural decision, are excluded by construction.

\textbf{Inference and issue scoring.} All reported inference runs use \textit{mini-swe-agent} inside isolated Harbor task environments. The agent receives the source repository checkout and the complete ordered sequence of typed prompts, but no reference answers or validation artifacts. User turns are delivered sequentially within one conversation. The agent may edit and test the repository during issue turns, and the harness records patch checkpoints. After the final turn, the source benchmark's \texttt{tests/test.sh} evaluates the cumulative patch in the same patched sandbox without resetting the repository. QA answers are stored separately and never enter the issue patch.

\textbf{QA validation.} Answer scoring uses GPT-Sol in Codex to compare the submitted answer with the hidden reference answer and its evidence chain. Original issue tests and creation-time proof execution provide deterministic evidence, but final QA correctness remains model-judged.

\textbf{Released dataset.} All generated tasks used in this paper are released as one public dataset on \hficon~Hugging Face, \textbf{\url{\hfdataseturl}}, over the same ordered 700-task SWE-Bench Pro subset; each experimental condition is a separate split. The \texttt{single\_prompt}, \texttt{split\_prompt}, and \texttt{split\_prompt\_concat} splits contain the prompt-only \textit{Single}, \textit{Split}, and \textit{Concat} conditions (see Appendix~\ref{app:experiments}). The remaining splits contain the QA-first/-last/-random placements and QA-only variants for the Terra and Sol question pools, plus the TFAS-optimized selection of Appendix~\ref{app:tfas}; every question ships with its reference answer (hidden from the evaluated agent), golden proof script, and creation-time proof-execution report. Content hashes and provenance manifests make every experimental condition exactly replayable. Alongside the ordered user prompts and validation artifacts, the dataset includes the inference trajectories of the evaluated agents from the pilot runs. Source repositories, gold patches, and verifier tests are not redistributed and remain with the original SWE-Bench Pro release~\citep{deng2025swebenchpro}.

\textbf{Released code.} The generation code is public on \codeicon~GitHub under Apache-2.0, \textbf{\url{\codeurl}}: the prompt-splitting and QA prompts, the deterministic QA ingestion, proof validation, and filtering steps, and the assembly of hybrid records. The Harbor orchestration that runs the QA agents inside task containers is not included.

\section{TFAS Details and Dataset Tuning}
\label{app:tfas}

TFAS compares three pairs of distributions: for each component $k\in\{L,T,R\}$---session-length bins, task types, and directed transitions---$P_k$ is the target distribution measured on long Production Sessions and $Q_k$ is the corresponding distribution over the benchmark's candidate trajectories. The score defined in Section~\ref{sec:taskflow} is
\begin{equation}
  S_k = 1-\operatorname{JSD}_2(P_k,Q_k), \qquad
  \operatorname{TFAS}=100\cdot\left(S_L \cdot S_T \cdot S_R\right)^{1/3},
  \label{eq:tfas}
\end{equation}
where, for two distributions $P$ and $Q$ on a shared support with $M=(P+Q)/2$,
\begin{equation}
  \operatorname{JSD}_2(P,Q)
  =\tfrac{1}{2}\operatorname{KL}_2(P\|M)
  +\tfrac{1}{2}\operatorname{KL}_2(Q\|M).
\end{equation}
We use length bins $1,2,3,4,5,6$--$10$, and $11+$. For the current target, bins below three have zero target mass and candidate trajectories below three turns serve as ablation baselines. The task distribution is defined over the 13 labels retained after excluding \textsc{greeting}, \textsc{meta-question}, and \textsc{other}, which are not instantiable as repository tasks. We remove those messages before constructing transitions, so an excluded turn between two retained messages is skipped and the retained messages become adjacent. The transition distribution contains the resulting ordered pairs and explicit \textsc{start}/\textsc{end} edges around each filtered sequence. This filtering applies only to TFAS; the Sankey diagrams in the paper are unfiltered and retain all 16 labels to show the source data transparently. Each of the three distributions is normalized independently. Jensen--Shannon divergence remains finite when one distribution assigns zero mass to an event, but rare transitions can still be unstable; bootstrap intervals should therefore resample complete sessions rather than individual edges.

For each verified source task, the generated candidate set includes the original single prompt, one to three issue turns formed by merging adjacent split prompts, zero to three available QA turns, and valid placements of the selected QA turns. Long-slice tuning filters this set to trajectories with at least three turns, then selects one candidate per source task to maximize TFAS against the long-session Production target. The component scores and full candidate-selection configuration should accompany every aggregate TFAS.

\textbf{TFAS-optimized selection.} 
The question pool contains 467 validated Sol-generated questions covering 255 of the 700 splittable SWE-Bench Pro tasks. A randomized local search (64 restarts, up to 20 sweeps, fixed seed) decides, per task, how many of its questions to keep and where to place them, maximizing Equation~\ref{eq:tfas} against the long-session Production target; the 445 tasks without questions keep their three issue turns. The optimizer retained 454 of the 467 questions, augmenting 113 sessions with one question, 85 with two, and 57 with three. The selected placements mix all patterns: 67 sessions carry their questions only before the issue turns, 86 only after, and 102 interleaved; by slot, 151 questions precede the first issue turn, 133 sit between issue turns, and 170 follow the last one. Table~\ref{tab:tfas-variants} reports the resulting alignment: adding QA lifts every component over pure \textit{Split}, fixed placements already reach TFAS 75.6--77.1, and optimizing count and placement adds mainly transition alignment, for a final TFAS of 77.8 versus 59.6 for \textit{Split} alone. The original single-turn prompts score $S_L{=}0$ against the three-message target (their length bin has zero target mass), which collapses the geometric mean to zero.

\begin{table}[h]
  \centering
  \caption{TFAS and its components against the long-session ($\geq$3 messages) Production target, computed on the same 700 splittable SWE-Bench Pro tasks. \textit{Split} keeps three issue turns; the QA rows place the Sol-generated questions at fixed positions; TFAS-optimized selects question count and placement per task.}
  \label{tab:tfas-variants}
  \begin{tabular}{lrrrr}
    \toprule
    Trajectory variant & $S_L$ & $S_T$ & $S_R$ & TFAS \\
    \midrule
    \textit{Split}               & 49.1 & 74.0 & 58.2 & 59.6 \\
    \textit{Split+QA-first}      & 83.9 & 81.5 & 63.7 & 75.8 \\
    \textit{Split+QA-last}       & 83.9 & 81.5 & 63.3 & 75.6 \\
    \textit{Split+QA-random}     & 83.9 & 81.5 & 67.0 & 77.1 \\
    \textit{TFAS-optimized}      & 84.0 & 81.5 & 68.8 & \textbf{77.8} \\
    \bottomrule
  \end{tabular}
\end{table}

Table~\ref{tab:tfas-swe-chat} additionally preserves an earlier fixed-variant ablation against SWE-chat (64 tasks, Terra-generated questions): calibrating to a different target favors different variants, which is precisely why the target must be stated.

\begin{table}[h]
  \centering
  \caption{TFAS ablation against SWE-chat, reported separately because it is not the paper's calibration target. All values are percentages on the same 64 source tasks.}
  \label{tab:tfas-swe-chat}
  \begin{tabular}{lrrrr}
    \toprule
    Condition & $S_L$ & $S_T$ & $S_R$ & TFAS \\
    \midrule
    \textit{Single}          & 37.0 & 52.9 & 17.3 & 32.4 \\
    \textit{Split}           & 25.4 & 52.6 & 37.7 & 36.9 \\
    \textit{Split+QA-first}  & 50.1 & 53.9 & 31.3 & 43.9 \\
    \textit{Split+QA-last}   & 50.1 & 53.9 & 32.5 & 44.5 \\
    \textit{Split+QA-random} & 50.1 & 53.9 & 35.1 & \textbf{45.6} \\
    \bottomrule
  \end{tabular}
\end{table}

\section{QA Answer Accuracy}
\label{app:qa-accuracy}

\begin{figure}[ht]
  \centering
  \includegraphics[width=0.9\linewidth]{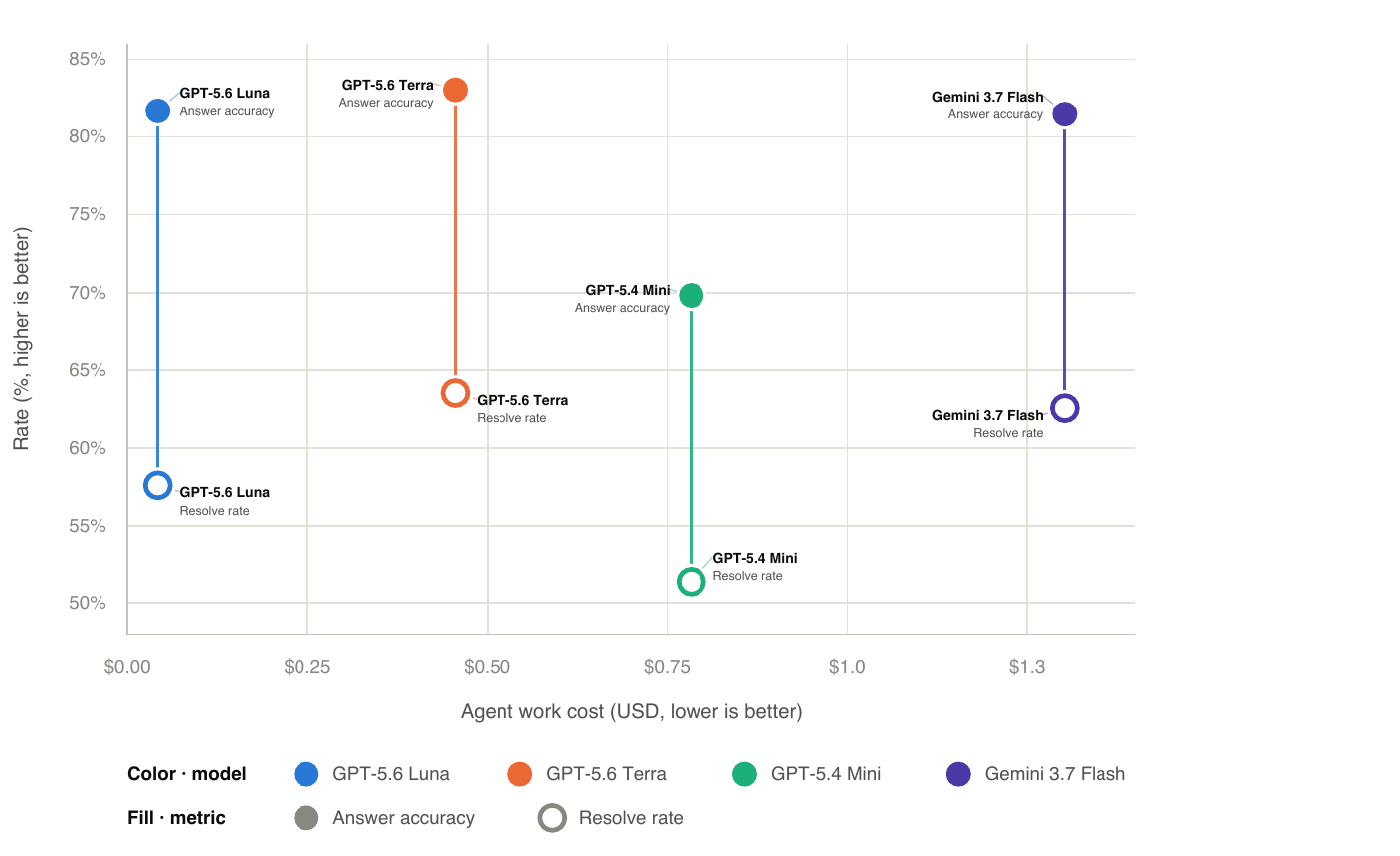}
  \caption{Answer accuracy (solid) and resolve rate (hollow) versus agent cost for the TFAS-optimized \textit{Split+QA} trajectories with Sol-generated questions; each vertical pair comes from the same runs of one model. Both metrics are computed on the 255 issues that carry QA questions, not on all 700 tasks, so resolve rate is not directly comparable to Figure~\ref{fig:pilot-metrics}.}
  \label{fig:qa-accuracy}
\end{figure}

At evaluation time the agent answers each repository question in text, and the answer is judged against the hidden reference (Appendix~\ref{app:pipeline}); \emph{answer accuracy} is the fraction of questions judged correct. Figure~\ref{fig:qa-accuracy} measures it on the target benchmark itself: the TFAS-optimized \textit{Split+QA} trajectories carrying the Sol-generated question pool (454 questions across 255 of the 700 tasks; Appendix~\ref{app:tfas}), evaluated with the same four models and runs as Figure~\ref{fig:pilot-metrics}. Each model contributes two points from the same trajectories---answer accuracy over its questions and resolve rate over the same QA-scored issues---so every trajectory yields both a patch outcome and a QA outcome. Everything in the figure is computed on these 255 issues that carry questions, not on the full 700-task set.

\plotdep{Answer accuracy spans 69.8\% (GPT-5.4 Mini) to 83.0\% (GPT-5.6 Terra) across the four models.} We report the two metrics side by side because they come from the same trajectories, not to claim any relationship between them.

\clearpage
\section{Additional Experimental Results}
\label{app:experiments}

\begin{figure}[p]
  \centering
  \includegraphics[width=0.9\linewidth]{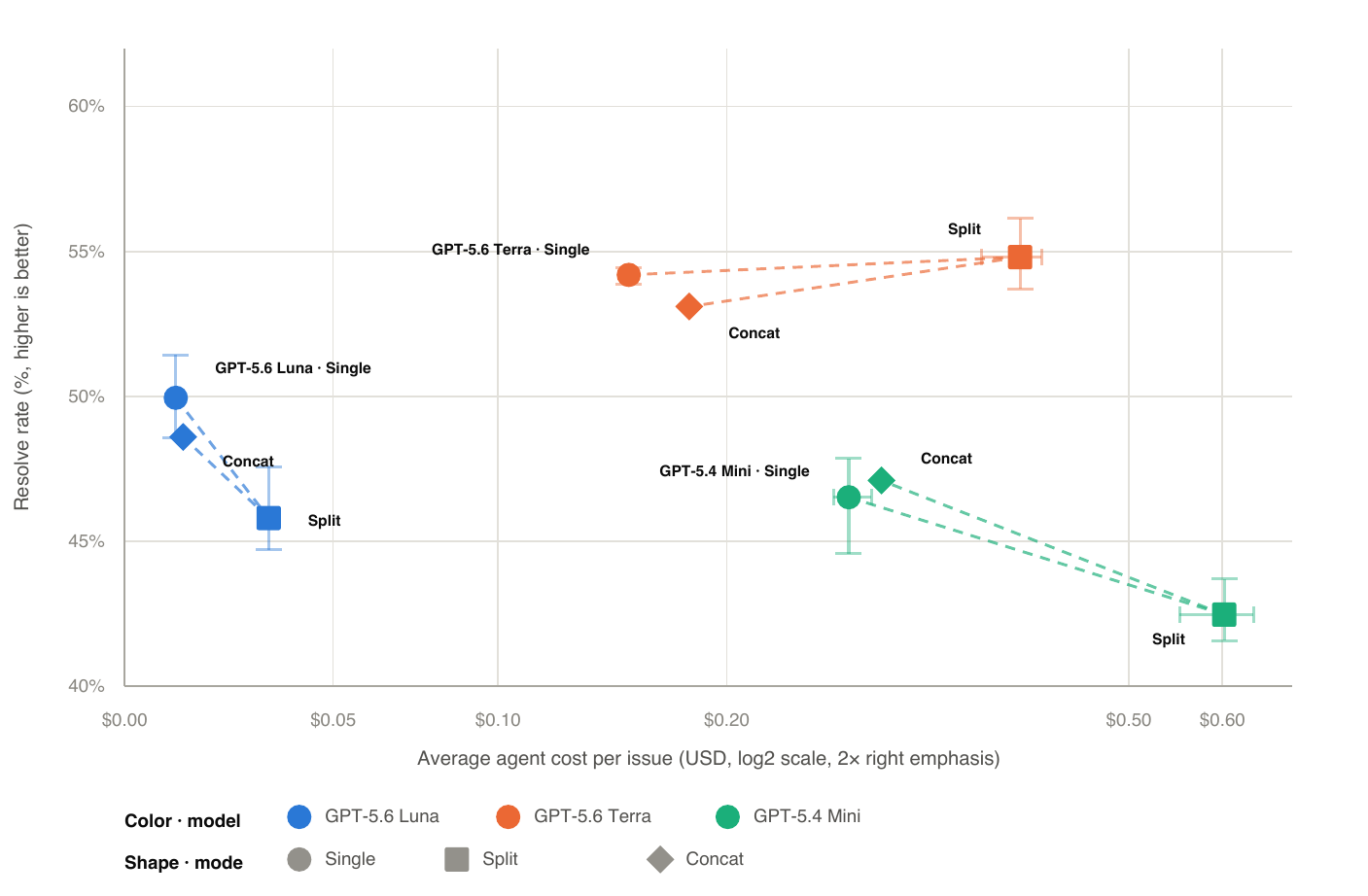}
  \caption{Resolve rate versus average agent cost per issue (log scale). \textit{Single} and \textit{Split} markers are arithmetic means of three runs on the 700 tasks; whiskers show observed min--max cost and resolve rate separately, not confidence intervals. Each \textit{Concat} marker is one 700-task run and therefore has no whiskers.}
  \label{fig:interaction-tax}
\end{figure}

\textbf{Protocol.} The models are GPT-5.6 Luna, GPT-5.6 Terra, and GPT-5.4 Mini, all run with \textit{mini-swe-agent} in Harbor and with reasoning effort set to medium; Figure~\ref{fig:pilot-metrics} additionally includes Gemini 3.7 Flash. Its \textit{Single}/\textit{Split}/\textit{Split+QA} comparison uses all 700 tasks and three runs per model and mode. \textit{Concat} delivers the three \textit{Split} segments in one message and has one available 700-task run per GPT model. Figure~\ref{fig:interaction-tax} therefore combines three-run summaries for \textit{Single} and \textit{Split} with single-run \textit{Concat} points; these comparisons are descriptive, not significance tests. Intermediate checkpoint results are reported below.

\textbf{Concat ablation (information preservation).} Automatic decomposition could lose or distort information from the original issue; if it did, any \textit{Split} effect would confound the interaction protocol with rewriting artifacts. The \emph{Concat} control removes this confound: it delivers the same three rewritten segments as \textit{Split}, but concatenated into a single message. The gap between \textit{Single} and \textit{Concat} therefore measures information change introduced by rewriting, while the gap between \textit{Concat} and \textit{Split} isolates the effect of sequential disclosure itself. Figure~\ref{fig:interaction-tax} shows all three conditions.

\begin{figure}[p]
  \centering
  \includegraphics[width=0.9\linewidth]{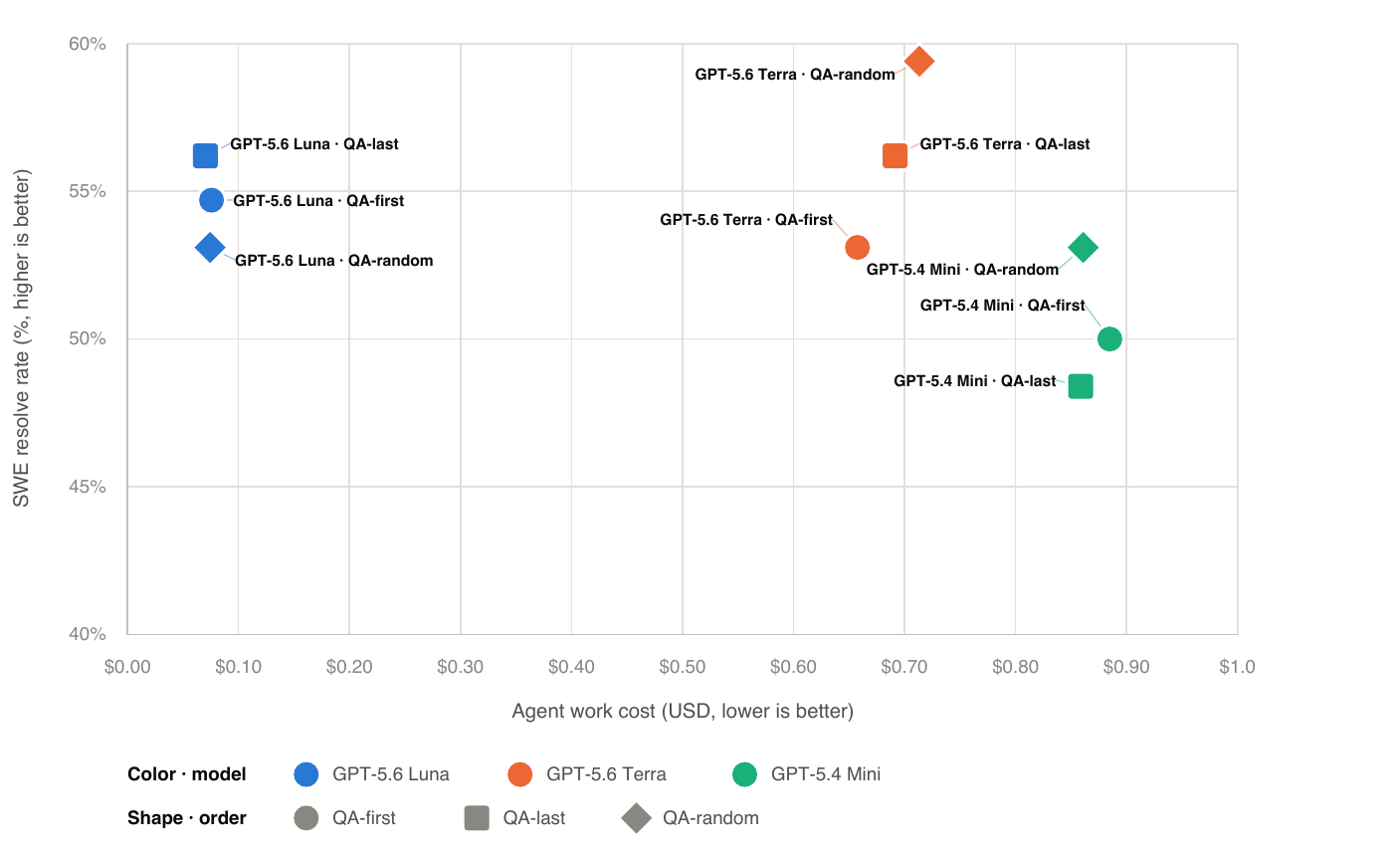}
  \caption{Resolve rate versus agent cost for QA-first (circle), QA-last (square), and QA-random (diamond) on the same 64 tasks and 114 questions; color encodes the model. Only the order of phases differs between the conditions.}
  \label{fig:qa-resolve}
\end{figure}

\begin{figure}[p]
  \centering
  \includegraphics[width=0.9\linewidth]{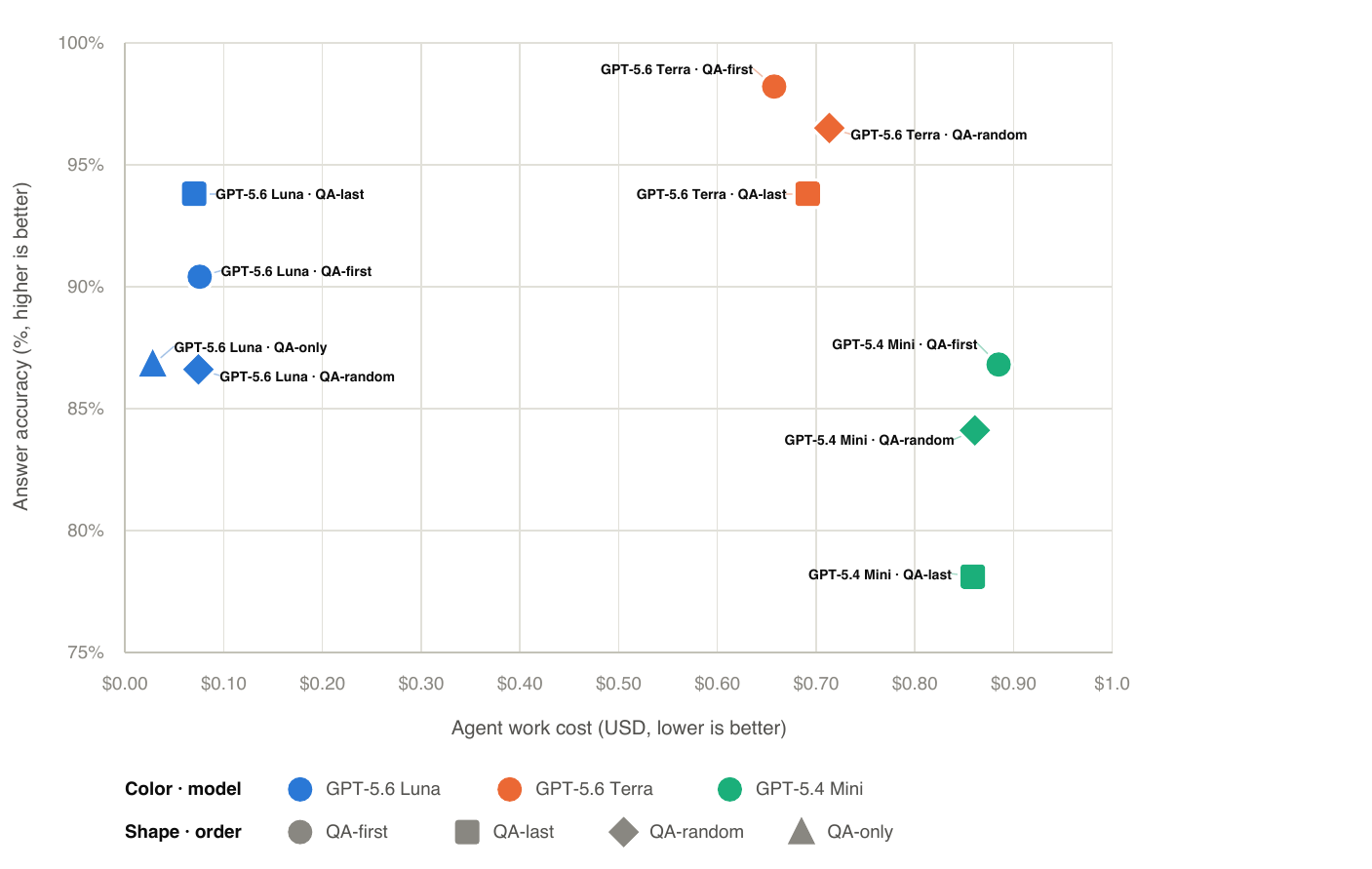}
  \caption{Answer accuracy versus agent cost per issue on the same 64 tasks and 114 Terra-generated questions. Color encodes the model, shape the placement of the QA turns; \textit{QA-only} delivers only the question turns, with no issue-resolve turns at all.}
  \label{fig:qa-accuracy-placement}
\end{figure}

\textbf{QA-placement ablation.} Placement shapes the transition distribution, and TFAS optimization moves questions to whatever positions the target Task Flow demands; before permitting that, we check whether position alone changes task difficulty. The placement policy of Section~\ref{sec:taskflow} admits three strategies for a session that combines issue turns and QA turns: \emph{QA-first} places all questions before the issue turns, \emph{QA-last} places them after, and \emph{QA-random} interleaves them at random positions. The ablation uses the same 64 instances and 114 Terra-generated questions in every condition; only phase order differs, so the comparison isolates ordering rather than the effect of adding QA (no matched no-QA run exists for these tasks). \plotdep{Resolve-rate differences between the strategies show no consistent ordering across models and are comparable to run-to-run variation} (Figure~\ref{fig:qa-resolve}). \plotdep{This is the license TFAS optimization relies on: within the tested range, moving questions around changes the measured Task Flow without changing how hard the underlying tasks are.} Answer accuracy under the same placements is 78.1--86.8\% for GPT-5.4 Mini, 86.6--93.8\% for GPT-5.6 Luna, and 93.8--98.2\% for GPT-5.6 Terra (Figure~\ref{fig:qa-accuracy-placement}). \plotdep{Placement shifts accuracy by a few points with no consistent direction across models, so we read question placement as having no real effect on answer accuracy either.} A \emph{QA-only} condition, in which the trajectory contains only the question turns and no issue turns at all, costs a fraction of any mixed trajectory and still reaches 86.8\% with Luna, matching its QA-random accuracy: the questions are answerable without any issue work, as the generation contract requires (every question must stay valid before and after the issue patch).

\begin{figure}[p]
  \centering
  \includegraphics[width=0.8\linewidth]{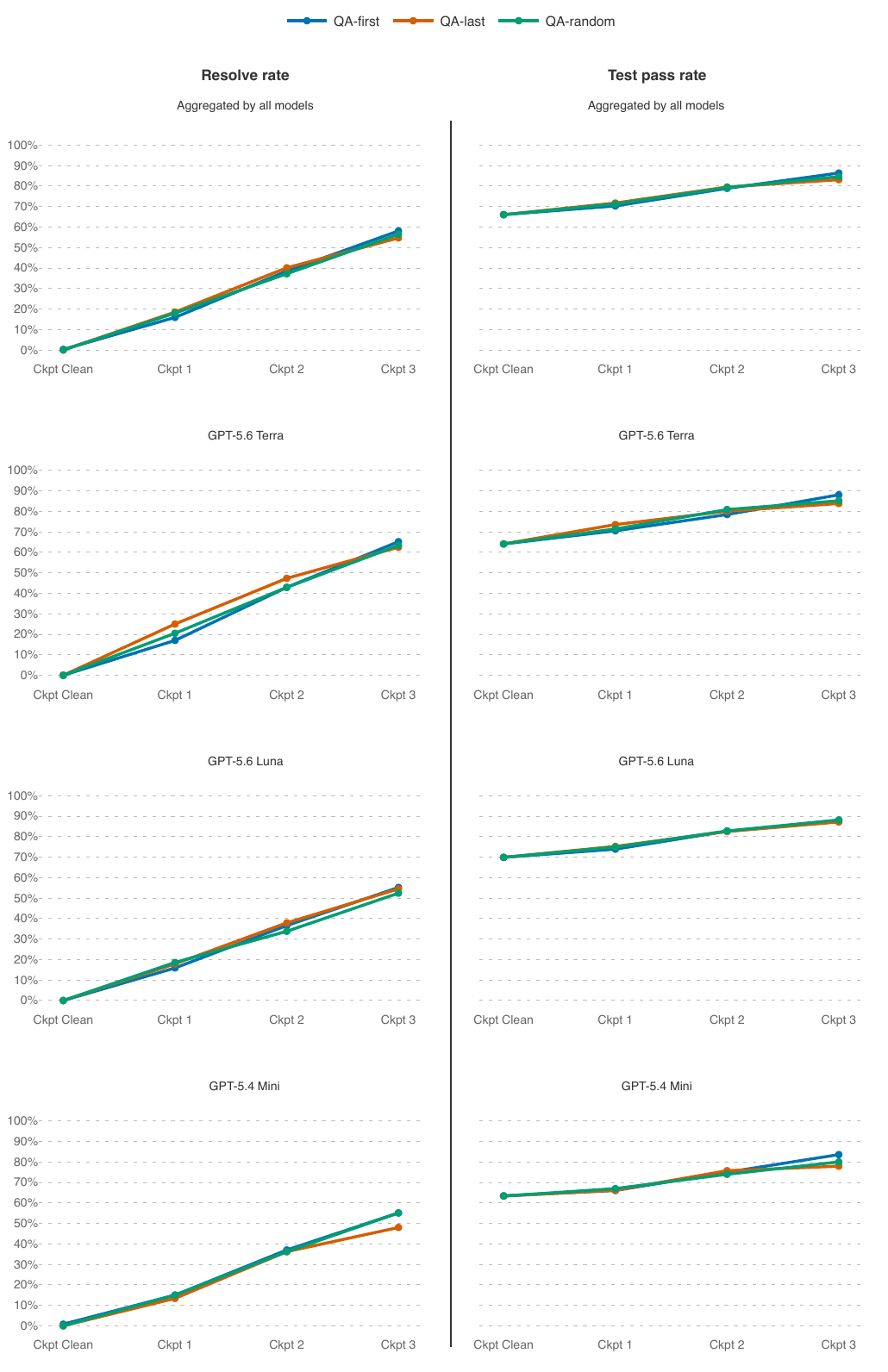}
  \caption{Checkpoint-level resolve rate (left) and test-pass rate (right) on the clean repository and after each of the three issue turns, under the three QA placements, aggregated and per model. QA turns do not modify the repository, so QA-last doubles as a \textit{Split}-only baseline. The placements are nearly indistinguishable at every checkpoint, while both metrics grow steadily with each issue turn.}
  \label{fig:checkpoint-signals}
\end{figure}

\textbf{Checkpoint-level signals.} \textit{Split} trajectories admit intermediate verification: after each issue turn, the harness snapshots the repository state and re-runs the original test suite, yielding a resolve rate and a test-pass rate at every checkpoint rather than a single terminal score. Figure~\ref{fig:checkpoint-signals} plots both metrics for the placement-ablation runs, measured on the clean repository and after the first, second, and third issue turns, aggregated and per model. Since QA turns never modify the repository, the QA-last arm doubles as a \textit{Split}-only baseline: all of its questions follow the issue turns, so its checkpoints measure pure splitting with no QA influence at all. At every checkpoint, for every model and in aggregate, the three placement curves lie nearly on top of each other: in this scenario, neither inserting repository questions nor their position measurably changes intermediate resolve or test-pass rates, while both metrics grow steadily as issue turns accumulate---the per-turn signal the interaction layer is designed to expose.

\clearpage
\section{Ethical Considerations and Broader Impacts}

Only aggregate statistics from Production Sessions are reported; the paper includes no session text or user identifiers, and the raw sessions are not released. The Task-Flow analysis is restricted to the long-session slice, while the full sample is used only for prevalence statistics. Public interaction corpora are cited and represented only through aggregate diagrams. The released task datasets (Appendix~\ref{app:pipeline}) contain only content derived from the public SWE-Bench Pro tasks and model-generated turns; no user data enters them. Better interaction-level evaluation could improve agent reliability and reveal the user effort hidden by terminal scores. Negative effects include additional inference cost, privacy risks from operational data, and the possibility that matching frequent behavior overweights noisy or low-value workflows. We mitigate these risks by reporting aggregates, preserving existing verified tasks, separating representative mixtures from targeted stress tests, and treating model-generated classifications as provisional.

\textbf{LLM use.} LLMs are core methodological components: they classify user-message intent, determine split eligibility, generate ordered prompt segments, extract repository facts, generate questions and reference answers, review candidates, and judge answers in the QA protocol. Executable task tests and creation-time proof execution provide non-textual evidence, but they do not remove model dependence from classification and generation.

\end{document}